\documentclass[runningheads]{llncs}

\usepackage{xcolor}
\definecolor{darkred}{rgb}{0.8,0,0.2}

\usepackage{svg}
\usepackage{hyperref}
\usepackage{listings}
\usepackage{parcolumns}
\usepackage{graphicx}
\begin{document}
\title{Explainable Machine Learning for liver transplantation}
%
%
\author{
Pedro Cabalar\inst{2}\orcidID{0000-0001-7440-0953} \and
Brais Muñiz\inst{1,2}\orcidID{0000-0002-9817-6666} \and
Gilberto Pérez\inst{1,2}\orcidID{000-0001-6169-6101} \and
Francisco Suárez\inst{3}\orcidID{0000-0002-9405-3538}
}
\authorrunning{Cabalar et al.}
%
\institute{
IRLab, CITIC Research Center \\ 
\and
University of A Coruña \email{\{cabalar;brais.mcastro;gilberto.pvega\}@udc.es}\\
\and
Digestive Service, Complexo Hospitalario Universitario de A Coruñna (CHUAC) \\
Instituto de Investigacióon Bioméedica de A Coruña (INIBIC) \\
University of A Coruña, SPAIN \email{francisco.suarez.lopez@sergas.es}
}

\maketitle              
\begin{abstract}
In this work, we present a flexible method for explaining, in human readable terms, the predictions made by decision trees used as decision support in liver transplantation.
The decision trees have been obtained through machine learning  applied on a dataset collected at the liver transplantation unit at the Coru\~na University Hospital Center and are used to predict long term (five years) survival after transplantation.
The method we propose is based on the representation of the decision tree as a set of rules in a logic program (LP) that is further annotated with text messages.
This logic program is then processed using the tool \texttt{xclingo} (based on Answer Set Programming) that allows building compound explanations depending on the annotation text and the rules effectively fired when a given input is provided.
We explore two alternative LP encodings:
one in which rules respect the tree structure (more convenient to reflect the learning process) and one where each rule corresponds to a (previously simplified) tree path (more readable for decision making).

\end{abstract}
%


\section{Introduction}


When Artificial Intelligence (AI) techniques are applied in a sensitive domain such as Healthcare, providing an explanation for the results is sometimes as important as the accuracy or correctness of those results, if not more.
Take, for instance, the liver transplantation domain and the problem of deciding a donor-receiver matching using a good prediction of the graft survival.
If significantly enough data are available, Machine Learning (ML) algorithms may obtain highly accurate predictions, but many ML techniques act as black boxes, failing to provide a verifiable explanation for their results.
This lack of explanations is especially problematic when the final decision may have critical consequences on the patients in the waiting list or even lead to legal implications.
Thus, most hospitals use the gravity of the patient's health as the only priority for the waiting list: the potential survival of the transplantation is just disregarded due to the lack of clear and fair rules that can be properly justified.

%
%
%
%

One ML technique that does not suffer from this black box limitation is \emph{decision tree} (DT) learning~\cite{Quin86}.
In DT, the result of the learning process is a tree whose nodes check conditions about the input features.
A prediction can be easily explained in human terms by following the corresponding path in the tree.
In the recent survey~\cite{connor20} of articles that apply AI to organ transplantation, more than a half of the approaches include a DT learning algorithm.
%
%
In the case of liver transplantation, Bertsimas et al.~\cite{Bertsimas2018}, used a big dataset with 1,618,966 observations to predict 3-month mortality or removal from the waiting list for a given patient.
ML techniques were applied to obtain two DT models: one for patients with hepatocellular carcinoma (HCC) and different one for the rest.
An interactive online tool\footnote{\url{http://www.opom.online/}} was built, including a simulator to make a prediction using 4 input variables from some patient's data.
The tool does not provide a specific explanation for the simulated prediction but, instead, it allows browsing the general DTs graphically, collapsing or expanding some parts of the tree.
In particular, the overwhelming width and detail of the non-HCC tree makes it very difficult to follow a certain path, even with the provided interactive browser.
So, strictly speaking, we can obtain an explanation for each prediction but, from a practical perspective, this approach lacks of the simplicity and ease of use that is expected from an explainable AI tool.

In this work, we present a flexible method for  explaining, in human readable terms, the 5-year survival predictions made by a DT trained on a dataset of liver transplantations.
The dataset was collected at the Digestive Service of the Coru\~na University Hospital Center (CHUAC), Spain.
The method consists in representing the DT as a logic program and using the tool \texttt{xclingo}~\cite{xclingo20} to annotate the program with natural language tags and construct the compound explanations.
%
%
We purpose two different translations into logic programming, each with its benefits and disadvantages, which will be discussed.
%
%

%


\section{An example of DT learning for predicting liver survival}
\label{sec:dataset_ml}

The dataset consisted of 258 transplants dating from 2009 to 2014 and each of those samples comprises 66 features both from the receiver and the donor.
As a target variable, we used the Boolean feature \texttt{goal\_death}, that points out if the 
patient died during the following 5 years after the surgery.
%
%
Numerical features were previously discretised using another DT, as described in \cite{ibm}.
The feature selection consisted in a chi-square test performed for each candidate feature against the target.
We took the 7 features with lowest $p$-value (i.e. highest significance for predicting the target variable) that are shown in Table~\ref{tab:pvalues} (prefixes ``don\_'' and ``rec\_'' respectively stand for donor and receiver).
%
%
%
%
\begin{table}[!htb]
    \begin{minipage}{.4\linewidth}
      \caption{Top 7 significant input features}
      \label{tab:pvalues}
      \centering
        \begin{tabular}{|l|c|}
        \hline
        Feature & P-value  \\
        \hline
        rec\_vhc & 0.015 \\
        rec\_afp & 0.042 \\
        rec\_abdominal\_surgery & 0.049 \\
        don\_microesteatosis & 0.082 \\
        rec\_hypertension & 0.111 \\
        rec\_provenance & 0.138 \\
        don\_acv & 0.146 \\
        \hline
        \end{tabular}
    \end{minipage}%
    \begin{minipage}{.6\linewidth}
      \centering
        \caption{Grid Search parameters}
        \label{tab:gridseach}
            \begin{tabular}{|l|c|}
            \hline
            Parameter & Possible values  \\
            \hline
            maximum depth & {5,9,11} \\
            splitting criterion & {entropy, gini-importance} \\
            maximum features & {$\sqrt{n\_features}$, $log_2(n\_features)$} \\
            \hline
            Best parameters & 9, entropy, $\sqrt{n\_features}$ \\
            \hline
            \end{tabular}
    \end{minipage} 
\end{table}
%
Due to the unbalanced distribution of the target class (death (0.76), alive(0.24)), a stratified splitting was used for dividing the data into train and test (75:25) while preserving the ratio of the target class.
Categorical features were label-encoded before training so that decision tree could process them.
The best parameters for training the decision tree were estimated by performing a stratified, 5-fold cross validation grid search over the training test.
Table~\ref{tab:gridseach} shows the parameter grid and the best parameters found.
%
%
The best parameters were used to train a DT that eventually produced an accuracy of 0.789 with a kappa value of 0.312.

%
%


\section{DTs as Explainable Logic Programs}
\label{sec:explanations}

Answer Set Programming (ASP)~\cite{brewka2011} is a declarative problem solving paradigm where problems are represented as a set of rules in a logic program and solutions to those problems are obtained in the form of answer sets.
ASP rules have the general form of ``\texttt{head :- body.}'' where both head and body are lists of \textit{literals}, that is, predicate atoms optionally preceded by \texttt{not}.
Intuitively, the rule head is derived as true if all the literals in the body are also true.
Rules with an empty body are called \textit{facts} and its head is always derived.
If we call the widely used ASP solver \texttt{clingo}~\cite{gekakaosscwa16a} on the following program:
\begin{flushleft}\ttfamily\small
holds(55,vhc,true). holds(55,don\_acv,true).\\
bad(P) :- holds(P,vhc,true), holds(P,don\_acv,true).
\end{flushleft}
we obtain an answer set including the two facts in the first line plus the atom {\tt bad(55)} derived from the rule in the second line.
\texttt{xclingo}~\cite{xclingo20} is an ASP tool built on top of \texttt{clingo} that explains why a given atom was derived by tracing the relevant fired rules.
To this aim, we may use textual descriptions to annotate specific rules (with the \texttt{\%!trace\_rule} directive) or any derivation of a given atom (with the \texttt{\%!trace} directive).
As an illustration, Listing~\ref{lst:xclingo_example} shows an annotated version of the previous example program and Listing~\ref{lst:xclingo_example_output}, the \texttt{xclingo} output.

The original obtained DT was automatically encoded into two different \texttt{xclingo} annotated programs: \texttt{nodes.lp} and \texttt{paths.lp}\footnote{All files publicly available in \url{https://github.com/bramucas/crystal-tree}}.
We also use two additional files: \texttt{extra.lp} which contains common code for \texttt{nodes.lp} and \texttt{paths.lp}; and \texttt{cases.lp} which contains the data from the transplant cases to be predicted.

%
%
%
%
%

\begin{minipage}{.52\textwidth}
\begin{lstlisting}[caption=\texttt{xclingo} annotated program.\label{lst:xclingo_example},frame=tlrb, basicstyle=\ttfamily\scriptsize, breaklines=true]{Name}
holds(55,vhc,true). 
holds(55,don_acv,true).
%!trace_rule {"Patient % may fail",P}
bad(P) :- holds(P,vhc,true), holds(P,don_acv,true).
%!trace {"% is %",F,V} holds(P,F,V).
%!show_trace bad(P).
\end{lstlisting}
\end{minipage}\hfill
\begin{minipage}{.4\textwidth}
\begin{lstlisting}[caption=\texttt{xclingo}'s explanation for \texttt{bad(55)}\label{lst:xclingo_example_output},frame=tlrb, basicstyle=\ttfamily\scriptsize, breaklines=true]{Name}
>> bad(55)  [1]
*
|__"Patient 55 may fail"
|  |__"rec_vhc is true"
|  |__"don_acv is true"
\end{lstlisting}
\end{minipage}




\begin{lstlisting}[caption=Fragment from \texttt{nodes.lp}.\label{lst:nodes},frame=tlrb, basicstyle=\ttfamily\scriptsize, breaklines=true]{Name}
%!trace_rule {"rec_hypertension is false"}
   tree_node(0,P,left) :- holds(P,rec_hypertension,false).
%!trace_rule {"rec_vhc is false"}
   tree_node(1,P,left) :- holds(P,rec_vhc,false), tree_node(0,P,left).
                          (...)
%!trace_rule {"rec_afp <= 1.84"}
   tree_node(6,P,left) :- le(P,rec_afp,184), tree_node(5,P,left).
alive(P) :- tree_node(6,P,left).
\end{lstlisting}
\noindent
\begin{minipage}{.45\textwidth}
\begin{lstlisting}[caption=Fragment from \texttt{paths.lp}.\label{lst:paths},frame=tlrb, basicstyle=\ttfamily\scriptsize, breaklines=true]{Name}
alive(P) :- 
 holds(P,rec_vhc,true), 
 holds(P,rec_abdominal_surgery,false), 
 holds(P,rec_hypertension,true), 
 le(P,rec_afp,20994), 
 le(P,don_microesteatosis,50).
\end{lstlisting}
\end{minipage}\hfill
\begin{minipage}{.52\textwidth}
\begin{lstlisting}[caption=Traces used by \texttt{paths.lp}. \label{lst:path-traces},frame=tlrb, basicstyle=\ttfamily\scriptsize, breaklines=true]{Name}
%!trace {"% is true",F} holds(P,F,true).
%!trace {"% is false",F} holds(P,F,false). 
%!trace {"% > %", F, T} gt(P,F,T).
%!trace {"% <= %", F, T} le(P,F,T).
%!trace {"% in (%,%]", F, Min, Max} between(P,F,Min,Max).
\end{lstlisting}
\end{minipage}
The \texttt{nodes.lp} program (partially shown in Listing~\ref{lst:nodes}) directly represents each DT edge using predicate \texttt{tree\_node(N,Patient,Dir)} where {\tt N} is the child node to be activated and {\tt Dir} its direction below the tree (left or right).
As we can see, each rule is annotated with a \texttt{\%!trace\_rule} describing the decision condition.
Leaves are encoded as rules with \texttt{alive(P)} or \texttt{not\_alive(P)}.
%
The following is an example of obtained explanation:
\begin{lstlisting}[frame=tlrb, basicstyle=\ttfamily\scriptsize, breaklines=true]{Name}
>> prediction(14)       [1]
  *
  |__"Bad (<5years)"
  |  |__"rec_afp > 509"
  |  |  |__"don_microesteatosis <= 50"
  |  |  |  |__"rec_afp <= 635"
  |  |  |  |  |__"rec_abdominal_surgery is false"
  |  |  |  |  |  |__"don_acv is true"
  |  |  |  |  |  |  |__"rec_afp <= 1244"
  |  |  |  |  |  |  |  |__"rec_vhc is true"
  |  |  |  |  |  |  |  |  |__"rec_hypertension is false"
\end{lstlisting}
whose cascade form reflects the order in which conditions are applied when traversing the tree, which comes from the (decreasingly) discriminatory power of each condition.
However, as a tree grows in depth, they become less readable and most discriminant features tend to be used repeatedly with different thresholds (as it happens above with \texttt{rec\_afp}) making the explanation less clear.

On the other hand, \texttt{paths.lp} (Listing~\ref{lst:paths}) just encodes a rule per each leaf in the original tree.
The head of the rule encodes the class of the leaf, and the body is a conjunction of all conditions traversed in the path.
%
An example of explanation from this second encoding would be:
\begin{lstlisting}[frame=tlrb, basicstyle=\ttfamily\scriptsize, breaklines=true]{Name}
>> prediction(14)       [1]
  *
  |__"Bad forecast (<5years)"
  |  |__"rec_abdominal_surgery is false"
  |  |__"don_acv is true"
  |  |__"rec_vhc is true"
  |  |__"rec_hypertension is false"
  |  |__"rec_afp in (509,635]"
  |  |__"don_microesteatosis <= 50"
\end{lstlisting}
%
\section{Conclusions}
As we can see, {\tt paths.lp} resulted in shorter explanations that are no longer in terms of the DT traversal.
In the example, we replaced the cascade of 8 conditions obtained before by a groups of 6 conditions (some of them in terms of intervals).
By grouping conditions, the explanation includes each feature used by the tree at most once, which guarantees readable explanations, even for the deepest paths.
Also, explanations can be easily adapted for different explanation needs (language, level of expertise, etc.) by modifying the text within {\tt trace} directives.
In our case, simpler and more general explanations for the patient could be provided while keeping all the detail in the explanations for the doctor.
%
%

As future work, we plan to improve the \texttt{paths.lp} by including probabilities extracted from the DT learning. 
%
We also plan to keep improving the accuracy of the obtained models by applying balancing techniques for the target feature.
Lastly, we also plan to continue collecting more transplant cases for the dataset.

%
%

%
%
%
\bibliographystyle{splncs04}
\bibliography{main}

\end{document}